\documentclass[letterpaper, 10pt, conference]{ieeeconf}   % Comment this line out
\usepackage{FG2021}

\FGfinalcopy % *** Uncomment this line for the final submission

\IEEEoverridecommandlockouts                              % This command is only
\usepackage{amsmath,graphicx}

\usepackage{amssymb}
\usepackage{mathtools}
\usepackage{graphics}
\usepackage{adjustbox}
\usepackage{stmaryrd}
\usepackage{tabu}
\usepackage{nth}
\usepackage{multirow}
\usepackage{bm}
\usepackage{csquotes}
\usepackage{url}
\usepackage{multicol}
\usepackage{lipsum}
\usepackage{mwe}
\usepackage[bottom]{footmisc}
\usepackage{hyperref}

\def\FGPaperID{132} % *** Enter the FG2021 Paper ID here

\title{\LARGE \bf
Leveraging Semantic Scene Characteristics and Multi-Stream Convolutional Architectures in a Contextual Approach for Video-Based Visual Emotion Recognition in the Wild
}

\author{\parbox{16cm}{\centering
    {\large Ioannis Pikoulis, Panagiotis P. Filntisis and Petros Maragos}\\
    {\normalsize
    School of ECE, National Technical University of Athens, Athens 15773, Greece}\\
    \vspace{0.09cm}\small{\texttt{\{el15198,filby\}@central.ntua.gr, maragos@cs.ntua.gr}}}
}

\begin{document}

\ifFGfinal
\thispagestyle{empty}
\pagestyle{empty}
\else
\author{Anonymous FG2021 submission\\\vspace{-0.2cm}Paper ID \FGPaperID }
\pagestyle{plain}
\fi
\maketitle

%%%%%%%%%%%%%%%%%%%%%%%%%%%%%%%%%%%%%%%%%%%%%%%%%%%%%%%%%%%%%%%%%%%%%%%%%%%%%%%%%%%%%%%%%%%%%%%%%%%%%%%%%%

\begin{abstract}

In this work we tackle the task of video-based visual emotion recognition in the wild. Standard methodologies that rely solely on the extraction of bodily and facial features often fall short of accurate emotion prediction in cases where the aforementioned sources of affective information are inaccessible due to head/body orientation, low resolution and poor illumination. We aspire to alleviate this problem by leveraging visual context in the form of scene characteristics and attributes, as part of a broader emotion recognition framework. Temporal Segment Networks (TSN) constitute the backbone of our proposed model. Apart from the \textit{RGB} input modality, we make use of dense \textit{Optical Flow}, following an intuitive multi-stream approach for a more effective encoding of motion. Furthermore, we shift our attention towards skeleton-based learning and leverage action-centric data as means of pre-training a Spatial-Temporal Graph Convolutional Network (\mbox{ST-GCN}) for the task of emotion recognition. Our extensive experiments on the challenging Body Language Dataset (BoLD) verify the superiority of our methods over existing approaches, while by properly incorporating all of the aforementioned modules in a network ensemble, we manage to surpass the previous best published recognition scores, by a large margin. 

\end{abstract}

%%%%%%%%%%%%%%%%%%%%%%%%%%%%%%%%%%%%%%%%%%%%%%%%%%%%%%%%%%%%%%%%%%%%%%%%%%%%%%%%%%%%%%%%%%%%%%%%%%%%%%%%%%

\section{INTRODUCTION}

The interpretation, perception and recognition of human affect has been the subject of rigorous studies and analyses across several scientific disciplines such as biology, psychology, sociology, neurology and last but not least, computer science. While the aforementioned cognitive sciences focus on the extraction of the available affective information, the fields of computer vision and machine learning aim at automating the recognition process through the development of novel techniques and algorithms which are capable of producing effective and robust encodings of such information.

The majority of past efforts in visual emotion recognition have been mostly limited to the analysis of facial expressions \cite{susskind2010toronto, goodfellow2013challenges, jung2015joint, mollahosseini2017affectnet, zhang2017facial}, while some studies have either incorporated information relative to body pose \cite{gunes2007bi, castellano2008emotion, banziger2012introducing, filntisis2019fusing} or have attempted to perform emotion recognition exclusively on the basis of body movements and gestures \cite{kapur2005gesture, karg2010recognition, piana2013set, saha2014study, sapinski2019emotion}. While some of these approaches perform well in certain specified settings, they fail to interpret real-world scenarios. 
%An example illustrating this point is the fact that emotion recognition systems often deal with instances of people whose facial features are fully visible and their body joints are unoccluded, something which does not generally conform to reality. 
This is because emotion recognition systems are, more often than not, expected to operate on instances of people whose facial features are fully visible and their body joints are unoccluded, something which does not generally conform to reality. 

Evidence from psychology related studies suggest that visual context, in addition to facial expression and body pose, provides important information to the perception of people’s emotions. Dudzik et al. \cite{dudzik2019context} propose two sources of context as means of interpreting emotional behavior, namely \textit{perceiver knowledge/experience} and \textit{perceivable \mbox{encoding} context}. 
Wieser and Brosch \cite{wieserbrosch} highlight situational context as the primary aspect of the latter, with features that mainly revolve around the visual scenes in which the depicted emotional behaviors are embedded. Barrett and Kensinger \cite{barrett2010context} report that the structural features of the face, when viewed in isolation, often prove to be insufficient for perceiving emotion. Furthermore, empirical findings suggest that the categorization of facial expressions is speeded up at the sight of congruent scenes \cite{righart2008recognition}, while both positive and negative contexts result in significantly different ratings of faces compared with those presented in neutral contexts \cite{mobbs2006kuleshov}.
%Wieser and Brosch \cite{wieserbrosch} state that features that correspond to this context category revolve around demographic information related to the gender of the emotional behaviors as well as the situations and scenes in which they are embedded. As far as demographic features are concerned, some examples might include age, cultural background, gender and occupation. Additionally, examples of situational features are the location, depicted scene and illumination. Empirical findings also suggest that the perceiver’s pre-existing knowledge and experiences have a significant impact on the way they decode and recognize affective states. This knowledge mainly includes racial stereotypes, affective associations, social norms, cultural values as well as their mental state, needs, goals and expertise.

In this work, we aim at extending the concept of context-based visual emotion recognition in the dynamic setting of video sequences. Our approach to the problem rests on the late fusion of Temporal Segment Networks (TSN) \cite{wang2016temporal} and a Spatial-Temporal Graph Convolutional Network \mbox{(ST-GCN)} \cite{yan2018}. 
%Both the TSN and ST-GCN constitute flexible and lightweight frameworks that have been primarily utilized in action recognition related literature.
We extend the original TSN framework by incorporating multiple input streams that encode bodily, contextual, facial and generic scene-related features, enhancing our model's joint understanding of emotion and the depicted surrounding environments. 
To the best of our knowledge, our approach is the first to explicitly infuse scene characteristics as well as make effective use of multi-stream optical flow in an emotion recognition process.
Extensive ablation experiments, based on the recently assembled and challenging Body Language Dataset (BoLD), are carried out so as to study the various contributions of our methods. 
%By combining our modified TSN and a properly pre-trained ST-GCN we manage to achieve significant improvements over the state-of-the-art techniques with relation to the Body Language Dataset (BoLD) \cite{Luo2019ARBEETA}. 

The remainder of the paper is structured as follows: Firstly, we provide an overview of the most notable related work in the domain of context and skeleton-based visual emotion recognition. Subsequently, we analyze our proposed model architecture. Next, we present our experimental results on the BoLD dataset, followed by conclusive remarks.

%%%%%%%%%%%%%%%%%%%%%%%%%%%%%%%%%%%%%%%%%%%%%%%%%%%%%%%%%%%%%%%%%%%%%%%%%%%%%%%%%%%%%%%%%%%%%%%%%%%%%%%%%%

\section{RELATED WORK}

Kosti et al. \cite{kosti2019context} made one of the first notable contributions towards context-based emotion recognition by introducing the EMOTIC dataset as well as providing a baseline model that was trained and evaluated on the latter. Their baseline model consisted of three modules: two ConvNet feature extractors (one for each of the body and context input streams) and one fusion network. A notable improvement in recognition performance over the baseline model came along the EmotiCon framework, as it was introduced by \mbox{Mittal et al.} \cite{mittal2020emoticon}. Their main contributions are associated with the incorporation of multiple modalities in the task of context-based emotion recognition, including the face, pose, inter-agent interactions and socio-dynamic context. %Additionally the information from the various modalities were combined through an early fusion scheme while imposing a modified cross-entropy-based loss function on each one of the separate modalities as well as the final predictions.

%Some works have also focused on extracting visual representations from images that preserve the semantic relations found in embeddings built from words. Wei et al. \cite{wei2020learning} trained a feature extraction network, EmotionNet, which they further regularized using joint text and visual embeddings. Their model achieved competitive zero-shot performance on the EMOTIC dataset against the original fully supervised baseline. 
Subsequently, researchers shifted their attention towards video-based emotion recognition in context. Lee et al. \cite{lee2019context} introduced the Context-Aware Emotion Recognition benchmark which is comprised of 13.201 TV video clips and a total of 1.1M frames. Moreover, a baseline model was proposed, 
%i.e. the CAER-Net, 
featuring a face and a context encoding stream which were merged using an adaptive-fusion network. 

After the recent assemble of the Body Language Dataset (BoLD), Luo et al. \cite{Luo2019ARBEETA} furthered their contributions by comparing various network configurations and finally providing a baseline model for the task of categorical and continuous emotion prediction. Among the examined methodologies were: motion-based descriptors, i.e. histograms of optical flow and motion boundary histograms, skeleton-based learning through an ST-GCN and Laban Movement Analysis \cite{Laban1950TheMO} as well as pixel-level learning through two-stream convolutional and TSN architectures. Filntisis et al. \cite{NTUA_BEEU} proposed the incorporation of a contextual feature encoding branch and the addition of visual-semantic embedding loss based on Global Vectors (GloVe) \cite{pennington2014glove} word embeddings, achieving state-of-the-art performance on BoLD.

As graph-based neural networks have proved to be powerful tools for determining human actions \cite{yan2018, Shi_2019_CVPR, Li_2019_CVPR, Si_2019_CVPR, Liu_2020_CVPR}, there have also been attempts to adapt them for the task of emotion recognition.
%Sapi{\'n}ski et al. \cite{sapinski2019emotion} analyzed motion capture data that were recorded using a Microsoft Kinect v2 sensor and proposed a sequential model of affective movement based on low level features inferred from the spacial location and the orientation of joints within the tracked skeletons. 
Bhattacharya et al. \cite{bhattacharya2020step} introduced a classifier network for the task of emotion recognition through gaits, as well as a realistic gait generator, with the ST-GCN architecture being the common denominator between the two. 
Sheng et al. \cite{sheng2021multi} proposed an Attention Enhanced Temporal Graph Convolutional Network (AT-GCN) as part of a multi-tasking framework which can jointly learn representations relative to both emotion and identity recognition.
%Moreover, \mbox{Shi et al.} \cite{shi2021skeleton} proposed a self-attention enhanced spatial-\mbox{temporal} graph convolutional network in which the spatial convolutional part modeled the skeletal structure of the body as a static graph, and the self-attention part dynamically constructed more connections between the joints, capturing supplementary affective information. 

%%%%%%%%%%%%%%%%%%%%%%%%%%%%%%%%%%%%%%%%%%%%%%%%%%%%%%%%%%%%%%%%%%%%%%%%%%%%%%%%%%%%%%%%%%%%%%%%%%%%%%%%%%

\section{MODEL ARCHITECTURE}

\begin{figure*}[htb]
\centering
\includegraphics[scale=0.5]{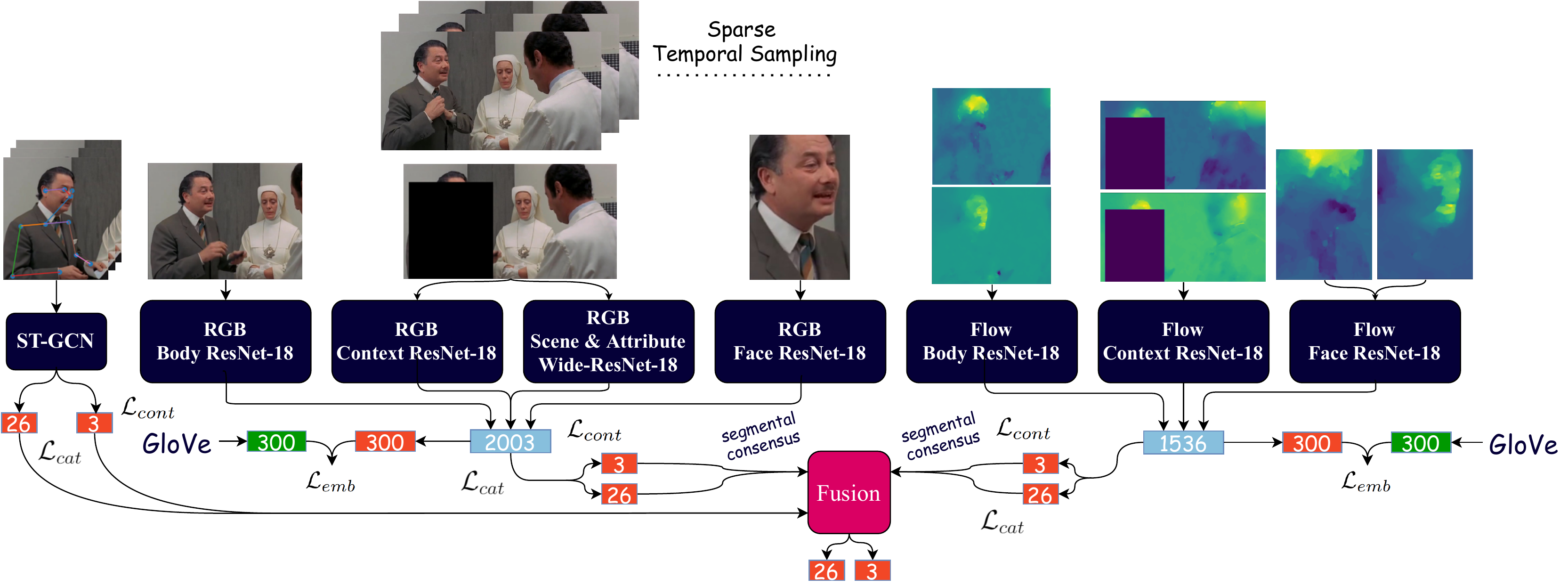}
\caption{Complete schematic diagram of the proposed network ensemble, featuring an ST-GCN module and three TSN input streams (\textit{body}, \textit{context}, \textit{face}) for both the \textit{RGB} and \textit{Optical Flow} modalities, plus a \textit{scene} \& \textit{attribute} related stream, especially for the \textit{RGB} modality.
Concatenated feature vectors are depicted in cyan, fully-connected layers are depicted in orange and GloVe word embeddings are depicted in green, along with their dimensionality or number of hidden units. The ST-GCN inherently produces video-level predictions, while in the case of the TSN-RGB and TSN-Flow, this requires the prior application of segmental consensus upon the corresponding snippet-level predictions (26 confidence scores for discrete emotions, 3 regressed values for VAD dimensions).  
Final predictions are obtained through late score fusion.} 
\label{fig:network}
\end{figure*}

A complete schematic diagram of our proposed network ensemble is shown in Fig. \ref{fig:network}. The backbone of our network implementation resides in a combination of the two-stream convolutional \cite{simonyan2014two} and TSN \cite{wang2016temporal} architectures, both of which were initially proposed for video-based action recognition. During TSN training, any given input video sequence $\mathcal{V}$ is firstly divided into $K$ segments $\{S_{1},...,S_{K}\}$ of equal durations. The TSN operates on a set of $K$ snippets, with each snippet constituting an instance that has been randomly sampled from the corresponding segment. More formally, the output of a temporal segment network is modeled as follows:

\vspace{-0.3cm}
\begin{equation}
\resizebox{.91\hsize}{!}{$\textrm{TSN}(T_{1},...,T_{K})=\mathcal{H}\Big(\mathcal{G}\big(\mathcal{F}(T_{1};\bm{\mathbf{W}}),\dots,\mathcal{F}(T_{K};\bm{\mathbf{W}})\big)\Big)$}
\end{equation}

\noindent where $\{T_{1},...,T_{K}\}$ denote the snippets, $\bm{\mathbf{W}}$ denotes the network trainable parameters, $\mathcal{F}$ denotes the snippet-level network predictions, $\mathcal{G}$ denotes a segmental consensus function and $\mathcal{H}$ denotes a prediction function.

Firstly, we will present the TSN structure regarding the \textit{RGB} modality, along with our proposed extensions for the enhancement of emotion understanding. Next, we will do the same for the \textit{Optical Flow} modality, and finally we will present the part of the architecture relative to skeleton-based learning. We choose to utilize 18-layer ResNets \cite{he2016deep} as our primary feature extractors for all the subsequent convolutional branches. ResNets constitute state-of-the-art ConvNet backbones, offering a valuable trade-off between performance and computational complexity. Moreover, the ResNet-18 variant produces 512-dim deep feature vector representations for each given input image.

%%%%%%%%%%%%%%%%%%%%%%%%%%%%%%%%%%%%%%%%%%%%%%%%%%%%%%%%%%%%%%%%%%%%%%%%%%%%%%%%%%%%%%%%%%%%%%%%%%%%%%%%%%

%%%%%%%%%%%%%%%%%%%%%%%%%%%%%%%%%%%%%%%%%%%%%%%%%%%%%%%%%%%%%%%%%%%%%%%%%%%%%%%%%%%%%%%%%%%%%%%%%%%%%%%%%%

\subsection{TSN-RGB}
\label{ssec:TSN-RGB}
\subsubsection{Body}
\label{sssec:rgb-body}
A single RGB image usually encodes static appearance at a specific point in time, lacking the contextual information about previous and next frames. We begin by training a standard TSN using the RGB modality and the body crops of each frame instance. For the calculation of the necessary body bounding boxes, we make use of the coordinates of 18 body joints that have been successfully tracked along the entirety of each video sequence and are being provided by the distributors of the dataset. The body branch feature extractor is pre-trained on ImageNet \cite{deng2009imagenet}. 
%Unless specified otherwise, all other ConvNet backbones that will later be used for other input streams, utilize the same architecture and only differ in the pre-training aspect.

\subsubsection{Context}
\label{sssec:rgb-context}
We incorporate a context stream in the form of RGB frames whose primary depicted agents have been masked out. For the acquisition of the masks we use the body bounding boxes that we have previously calculated and multiply them element-wise with a constant value of zero. Contextual feature extraction is a scene-centric task, and therefore we choose to initialize the corresponding \mbox{ConvNet} backbone using the Places365-Standard \cite{zhou2016places}, a large-scale database of photographs, labeled with scene \mbox{semantic} \mbox{categories}. 

\subsubsection{Face}
\label{sssec:rgb-face}
We introduce an input stream which explicitly operates on extracted face crops. For the localization and extraction of faces we use the five body joints that correspond to the eyes, ears and nose of each depicted instance. These joints are used to calculate the smallest bounding box that contains the head of the agent. As the pose of an agent might result in partial or complete occlusion of their facial features, the successful extraction of the face region is not guaranteed. The \mbox{ConvNet} backbone of the face branch receives manual pre-training on AffectNet \cite{mollahosseini2017affectnet} which constitutes the largest facial expression database, containing over 1M images, annotated on both categorical and dimensional level.

\subsubsection{Scenes and Attributes}
\label{sssec:rgb-scene&attributes}
The depicted scene along with its attributes hold valuable information relative to human emotion understanding, especially in cases where primary sources of affective information, such the face and body, are occluded. Therefore, we aspire to enrich our model's perception of context by directly extracting the Places365 scene-specific scores and the corresponding Scene UNderstanding (SUN) \cite{patterson2012sun} attributes through an 18-layer Wide-ResNet \cite{zagoruyko2016wide} which has been jointly pre-trained on both of the aforementioned databases. 

The Places \cite{zhou2016places} database is a quasi-exhaustive repository of 10M scene photographs, labeled with 476 scene semantic categories. We use only a subset of the latter, namely the Places365-Standard which features 1.8M images and 365 scene categories. Moreover, the SUN attribute database \cite{patterson2012sun} constitutes a subset of the SUN categorical database \cite{xiao2010sun}, comprised of 14,000 images that are annotated using a taxonomy of 102 scene attributes. Some of the categories that are included in the Places365 dataset are: amusement park, basketball court, cemetery, jail, cell, lecture room, museum, office, sauna, soccer field, etc. Some of the scene attributes included in the SUN dataset are: competing, socializing, working, exercise, praying, open-area, enclosed-area, stressful, etc. It is quite evident that the environment and scene depicted in an image can be closely related with the emotions of the people that are present. For example, an image of a funeral that is located at a cemetery, suggests a strong correlation between the above oppressive setting and the generally negative and sad feelings shared among the depicted people. Provided that our model is capable of leveraging the hinted correlations, incorporating scene specific information can potentially boost its overall emotion recognition performance.

Given an input image, the feature extractor, through each last convolutional block produces feature maps \mbox{$\bm{\mathbf{Z}}\in\mathbb{R}^{512\times14\times14}$}. After the application of an average pooling layer, a deep feature vector representation is formed and fed into a FC layer with weights      $\bm{\mathbf{W}}_{\textrm{scenes}}\in\mathbb{R}^{512\times365}$, producing class confidence scores $\hat{\bm{\mathbf{y}}}_{\textrm{scenes}}\in\mathbb{R}^{1\times365}$. An additional set of pre-trained weights, namely $\bm{\mathbf{W}}_{\textrm{attr}}\in\mathbb{R}^{512\times102}$ can be used for the prediction of confidence scores $\hat{\bm{\mathbf{y}}}_{\textrm{attr}}\in\mathbb{R}^{1\times102}$ for 102 scene attributes that are included in the SUN dataset. The corresponding scene and attribute classification probabilities are calculated after the row-wise application of the softmax function. Subsequently, the produced scene and attribute probability scores are concatenated with the extracted deep features from all the aforementioned input streams. After the initialization of the feature extractor with the aforementioned pre-trained models, weight parameters are kept frozen during the training phase. 

The inclusion of all input streams of the TSN-RGB results in a 2003-dim concatenated feature vector.

\subsubsection{Loss Functions}
\label{sssec:loss}
For the training of the continuous emotion prediction branch, we use a standard \textit{mean squared error} (MSE) loss $\mathcal{L}_{\textrm{cont}}$ along the three emotional dimensions of \textit{valence}, \textit{arousal} and \textit{dominance}. As far as categorical emotion prediction is concerned, the ground truth targets are provided in the form of confidence scores. Firstly, we apply a sigmoid function to the barebones extracted class scores and then impose an MSE loss between the predicted and ground truth confidence scores. We denote this loss term as $\mathcal{L}_{\textrm{{cat}}_{1}}$. Secondly, after binarizing the ground truth confidence scores with a threshold of 0.5, we apply a binary cross-entropy loss between the produced and ground truth multi-hot target labels. We denote this term as $\mathcal{L}_{\textrm{cat}_{2}}$. We also enforce semantic congruity between the extracted visual embeddings and the categorical label word embeddings from a 300-dim GloVe \cite{pennington2014glove} model, pre-trained on Wikipedia and Gigaword 5 data, in the same manner as in \cite{NTUA_BEEU}. More specifically, given an input image $\bm{\mathbf{X}}$, we transform the concatenated visual embeddings $f_{v}(\bm{\mathbf{X}})$ into the same dimensionality as the word embeddings $f_{t}(y^{i})$ through a linear transformation $\bm{\mathbf{W}}_{\textrm{emb}}$, with $\bm{\mathbf{y}}$ being the corresponding multi-hot target vector and $i$ being the categorical label class index. We later apply an MSE loss between the transformed visual embeddings and the average of word embeddings that correspond only to the positive labels of the given ground truth target vector and denote this term as $\mathcal{L}_{\textrm{emb}}$:
\begin{equation}
    \mathcal{L}_{\textrm{emb}}=\|\bm{\mathbf{W}}_{\textrm{emb}}f_{v}(\bm{\mathbf{X}})-\frac{1}{|\mathcal{P}|}\sum_{y^{i}\in{\mathcal{P}}}f_{t}(y^{i})\|_{2}^{2}
\end{equation}
\noindent where $\mathcal{P}$ denotes the set of positive class labels for a given target vector $\bm{\mathbf{y}}$ and $|\mathcal{P}|$ denotes the cardinality of that set. The whole network can be trained in an end-to-end manner by minimizing the combined loss function:                            

\begin{equation}\label{losses}
    \mathcal{L}=\mathcal{L}_{\textrm{cat}_{1}}+\mathcal{L}_{\textrm{cat}_{2}}+\mathcal{L}_{\textrm{cont}}+\mathcal{L}_{\textrm{emb}}
\end{equation}

\subsection{TSN-Flow}
\label{ssec:TSN-Flow}

Similarly to the case of temporal stream ConvNets in the original two-stream convolutional architecture \cite{simonyan2014two}, we experiment with training a TSN on stacked optical flow fields. Optical flow extraction is carried out via the TVL1 algorithm \cite{zach2007duality}. This form of dense optical flow is known to effectively encode motion between consecutive frames. We denote this model as TSN-Flow.

In all our subsequent implementations, we stack bidirectional optical flow fields from $L=5$ consecutive frames for each snippet. After decomposing each displacement vector into its horizontal and vertical components, we end up with a 10-channel input volume per segment, per input stream. To begin with, we train a standard TSN using the \textit{Optical Flow} modality and the body crops of each frame instance. The usage of body joint coordinates for the localization and extraction of the necessary bounding boxes remains the same as in the case of the RGB modality. 
Body-oriented dense optical flow encodes the movement of the primary agent depicted in each instance. Additionally, we incorporate a context stream, in the form of stacked optical flow fields whose primary depicted agents have been masked out. The context input stream effectively encodes the motion of any occasional secondary agent or object. Lastly, we introduce an input stream that focuses solely on the head and face movements of the primary agent. This is achieved by training an additional temporal ConvNet on small fragments of dense flow that correspond to the head region of each agent.

The features extracted using optical flow streams have distributions that greatly differ from their RGB counterparts. As optical flow values are discretized in the interval $[0,255]$, therefore sharing the same range with RGB images, we use RGB models to initialize the parameters of the temporal ConvNets. The feature extractors for all TSN-Flow streams were pre-trained on ImageNet \cite{deng2009imagenet}. Consequently, the weights of the first convolutional layer are modified so as to handle the input of optical flow fields. More specifically, the weights are averaged across the RGB channels and replicated by the number of channels of the temporal stream inputs.

The inclusion of all the aforementioned input streams results in a 1536-dim concatenated feature vector per input volume. During training we employ the same combined loss function as the one used for its RGB counterpart.

\subsection{Skeleton-Based Learning}
\label{ssec:skeleton-based}
As for the final source of affective information, we shift our attention towards the \textit{Human Skeleton} and attempt to incorporate a Spatial-Temporal Graph Convolutional Network (ST-GCN) \cite{yan2018}, as it was originally proposed for skeleton-based action recognition. We choose to deploy the vanilla ST-GCN consisting of 9 layers of spatial-temporal graph convolution operators (ST-GCN units). The features extracted from the last ST-GCN unit undergo average pooling and with the use of $1\times1$ convolutions, the final predictions are produced. 

\subsubsection{Joint Labeling Strategies}
The main variable setting of the ST-GCN configurations is the joint labeling strategy that is being used during the construction of the graph adjacency matrix, namely $\textit{uniform}$, $\textit{distance}$ or $\textit{spatial}$. With $\textit{uniform}$ being the simplest labeling strategy, all joints that are connected through a limb belong in the same subset, resulting in $K_v=1$ total subsets. The $\textit{distance}$ labeling strategy extends the concept of neighboring joints, as pairs of joints that are connected through a sequence of limbs are also taken into consideration, leading to a total of $K_v=D+1$ subsets, where $D$ is the maximum allowed distance between two neighboring joints (we choose $D=1$ for simplicity). Lastly, according to the $\textit{spatial}$ labeling strategy, neighboring joints are distinguished based on their individual distances from a fixed root (the neck), resulting in $K_v=3$ subsets.  

\subsubsection{Forward Propagation}
In the spatial-temporal case, the input feature map $\bm{\mathbf{H}}_{\textrm{in}}$ of a ST-GCN unit is represented as a tensor of shape $(C_{\textrm{in}},T_{\textrm{in}},V)$, where $C_{\textrm{in}}$ denotes the number of input channels, $T_{\textrm{in}}$ denotes the number of frames in the skeleton sequence and $V$ denotes the number of nodes. Firstly, the input tensor undergoes a \mbox{$(K_v\cdot C_{\textrm{out}})\times1\times1$} spatial graph convolution operation, with $C_{\textrm{out}}$ being the desired number of output channels and $K_v$ being the number of joint subsets that are formed based on the chosen labeling strategy. The resulting tensor is reshaped into $(K_v,C_{\textrm{out}},T_{\textrm{in}},V)$ and multiplied with the normalized adjacency matrix \mbox{$\bm{\mathbf{D}}^{-\frac{1}{2}}\hat{\bm{\mathbf{A}}}\bm{\mathbf{D}}^{-\frac{1}{2}}$}, where $\hat{\bm{\mathbf{A}}}=\mathbb{I}+\bm{\mathbf{A}}$ ($\mathbb{I}$ denotes the identity matrix) and $\bm{\mathbf{D}}$ is a diagonal matrix with elements \mbox{$D^{ii}=\sum_{j}\hat{A}^{ij}$}. In case of the distance and spatial partitioning strategies $(K_v>1)$, the adjacency matrix $\bm{\mathbf{A}}$ is formed by stacking $K_v$ matrices $\bm{\mathbf{A}}_{k}$, with each one corresponding to one of the $K_v$ joint subsets. If we ignore interlayer nonlinearities, then the aforementioned spatial convolution operation is equivalent to the original GCN \cite{kipf2017semisupervised} formula:

\begin{equation}\label{gcn_formula}
    \bm{\mathbf{H}}_{\textrm{out}}=\sum_{k}\bm{\mathbf{W}}_{k}\bm{\mathbf{H}}_{\textrm{in}}\bm{\mathbf{D}}_{k}^{-\frac{1}{2}}\hat{\bm{\mathbf{A}}}_{k}\bm{\mathbf{D}}_{k}^{-\frac{1}{2}}
\end{equation}

\noindent where $\bm{\mathbf{W}}_{k}$ are $C_{\textrm{out}}\times C_{\textrm{in}}\times1\times1$ weight matrices (the multiplication is replicated $T_{\textrm{in}}$ times in the temporal dimension and $V$ times in the spatial dimension). 
\mbox{$D_{k}^{ii}=\sum_{j}\hat{A}_{k}^{ij}+\alpha$} is the normalized diagonal matrix and $\alpha$ is set to 0.001 to avoid empty rows.
Additionally, learnable edge importance weighting can be implemented simply by multiplying element-wise the adjacency matrices $\hat{\bm{\mathbf{A}}}_{k}$ of Eq. \ref{gcn_formula} with a weight mask $\bm{\mathbf{M}}$, namely $\hat{\bm{\mathbf{A}}}_{k}\odot\bm{\mathbf{M}}$. The output feature map resulting from the spatial graph convolution undergoes a $C_{\textrm{out}}\times\Gamma\times1$ temporal convolution, with $\Gamma$ denoting the temporal kernel size, completing the processing pipeline of a ST-GCN unit.

Training in the case of the ST-GCN is driven by a combined loss function similar to the one described in Eq. \ref{losses}, with the sole difference being the exclusion of the categorical label embedding loss $\mathcal{L}_{\textrm{emb}}$.  

\subsubsection{Data Augmentation}
Joint coordinates are normalized in the range $[0,1]$ using the largest joint bounding box within each sequence. We then strictly follow the proposed methodologies of \cite{yan2018}. Firstly, we find the maximum sequence length $T$ within our dataset and pad every clip with zeros until it reaches that specified length. During the training phase, padding is applied randomly within the sequence while during inference the paddings are placed always at the end for consistency. Additionally, during training we perform random affine transformations on the skeleton sequences of all frames, with the aim of simulating camera movement.

After augmentation, input data is represented by tensors of size $(C,T,V)$. For each frame of a sequence, BoLD provides 18 tuples that contain 2D joint coordinates plus a detection confidence score associated with each joint, therefore in our case $C=3$ and $V=18$. In order to further reduce the effect of over-fitting, we also pre-train the \mbox{ST-GCN} on the Kinetics dataset \cite{kay2017kinetics} which has been extensively used for skeleton-based action recognition. %In that case, joint coordinates are centralized in the range $[-0.5, 0.5]$.

%%%%%%%%%%%%%%%%%%%%%%%%%%%%%%%%%%%%%%%%%%%%%%%%%%%%%%%%%%%%%%%%%%%%%%%%%%%%%%%%%%%%%%%%%%%%%%%%%%%%%%%%%%

\section{EXPERIMENTAL EVALUATIONS}
%The current section is dedicated to presenting the BoLD database as well as the extensive experiments which we have conducted on it.
\subsection{Dataset}

All of our upcoming experimental results are based on the standard train, validation and test splits of the Body Language Dataset (BoLD)\footnote{\url{https://cydar.ist.psu.edu/emotionchallenge}} which was assembled by \mbox{Luo et al. \cite{Luo2019ARBEETA}} and constitutes a database that focuses on bodily expressions of emotion. BoLD is comprised of 9,876 movie video clips of body movements, depicting a total of 13,239 human characters. 
The annotation of the dataset was performed using a crowdsourcing pipeline based on the Amazon Mechanical Turk. Instances are annotated in both categorical and dimensional level, utilizing the 26 emotional categories of the EMOTIC \cite{kosti2019context} dataset and the VAD \cite{mehrabianrussell77} dimensional model, respectively.   

As far as evaluation metrics are concerned, for categorical emotion prediction, \textit{average precision} (AP), i.e. the area under the precision-recall curve as well as the \textit{area under the receiver operating characteristic} (ROC-AUC) are used. For continuous emotion regression along the VAD dimensions, the coefficient of determination ($R^{2})$ is used. Performance comparison among different models is carried out on the basis of an aggregatory \textit{emotion recognition score} (ERS) which is calculated as follows:

\begin{equation}
    \textrm{ERS}=\frac{1}{2}\big(\textrm{m}R^{2}+\frac{1}{2}(\textrm{mAP}+\textrm{mRA})\big)
\end{equation}

\noindent where m$R^{2}$ denotes the mean coefficient of determination along the VAD dimensions while mAP and mRA denote the \textit{mean average precision} and the mean ROC-AUC over the 26 emotion categories respectively. 

%%%%%%%%%%%%%%%%%%%%%%%%%%%%%%%%%%%%%%%%%%%%%%%%%%%%%%%%%%%%%%%%%%%%%%%%%%%%%%%%%%%%%%%%%%%%%%%%%%%%%%%%%%

\begin{figure*}[t!]
\centering
\includegraphics[scale=0.26]{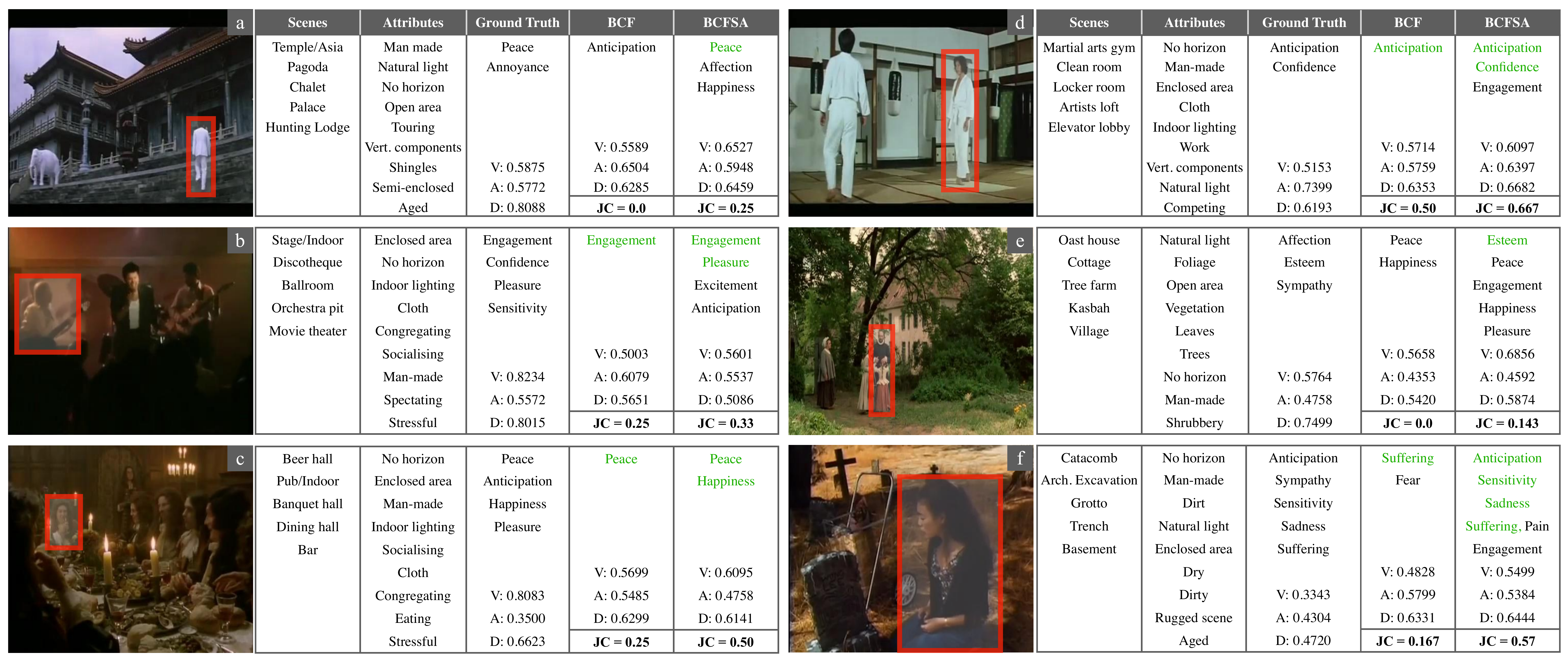}
\caption{Top-5 predicted scene categories, top-9 predicted attibutes, ground truth and predicted (regressed) emotion categories (VAD values) as well as \textit{Jaccard similarity coefficient} (\textrm{JC}) on samples that have been randomly selected from the BoLD validation set. All predictions are made at video level.} 
\label{fig:scenes_attrib}
\end{figure*}

%%%%%%%%%%%%%%%%%%%%%%%%%%%%%%%%%%%%%%%%%%%%%%%%%%%%%%%%%%%%%%%%%%%%%%%%%%%%%%%%%%%%%%%%%%%%%%%%%%%%%%%%%%

\subsection{Configuration Details}
\label{ssec:configuration}

For all experiments regarding TSN configurations, we use $K_{\textrm{train}}=3$ segments during training and $K_{\textrm{val}}=25$ segments during validation while the segmental consensus function has been chosen to be average pooling. Both the TSNs and \mbox{ST-GCN} are trained for 25 epochs with a batch size of 16, using the SGD optimizer, momentum equal to 0.9 and weight decay equal to ${10}^{-5}$. The initial learning rate is set to ${10}^{-3}$ in the case of TSNs and $5\cdot{10}^{-3}$ in the case of the ST-GCN. The learning rates are reduced by a factor of 0.1 whenever the monitored loss on the validation set plateaus. No particular data augmentation technique was applied for either the TSN-RGB or TSN-Flow, as the built-in variations of the BoLD dataset proved sufficient for avoiding over-fitting.

Apart from the previously described methodologies, we experiment with the partial batch-normalization scheme, or $\textit{Partial BN}$ for short, as proposed in \cite{wang2016temporal}. More specifically, after the initialization with pre-trained models, in every ConvNet feature extractor, we freeze the mean and variance of parameters of all batch normalization layers, except for the first one. This method is expected to work especially well in the case of temporal ConvNets and reduce the effect of over-fitting. More specifically, as the distribution of optical flow is different from the RGB images, the distribution of activation values in the first convolutional layer will also differ from the ones inherited through their initialization with RGB pre-trained models. 

All of our results were generated on a single NVIDIA GeForce RTX 2080 Ti GPU. All the code was
implemented using PyTorch\footnote{\url{https://pytorch.org/}}.

%%%%%%%%%%%%%%%%%%%%%%%%%%%%%%%%%%%%%%%%%%%%%%%%%%%%%%%%%%%%%%%%%%%%%%%%%%%%%%%%%%%%%%%%%%%%%%%%%%%%%%%%%%

\begin{table}[b!]
\centering
\caption{Performance comparison of various TSN-RGB model configurations on the BoLD validation set.}
\resizebox{8.6cm}{!}{
\begin{tabular}{|c|c|c|c|c|c|c|c|}
\hline
\multirow{2}{*}{Model}   & \multirow{2}{*}{Features} & \multirow{2}{*}{$\mathcal{L}_{\textrm{emb}}$} & \multirow{2}{*}{Partial BN} & Regression      & \multicolumn{2}{c|}{Classification} & \multirow{2}{*}{ERS} \\ \cline{5-7}
                         &                           &                       &                             & m$R^{2}$\mbox{}$\uparrow$           & mAP$\uparrow$               & mRA$\uparrow$               &                      \\ \hline
\multirow{8}{*}{TSN-RGB} & B                         & \multirow{5}{*}{No}   & \multirow{5}{*}{No}         & 0.0300          & 0.1419           & 0.5910           & 0.1983               \\ \cline{2-2} \cline{5-8} 
                         & BC                        &                       &                             & 0.0362          & 0.1468           & 0.6021           & 0.2053               \\ \cline{2-2} \cline{5-8} 
                         & BCF                       &                       &                             & 0.0531          & 0.1615           & 0.6213           & 0.2222              \\ \cline{2-2} \cline{5-8} 
                         & BCFS                      &                       &                             & 0.0597          & 0.1736           & 0.6428           & 0.2347               \\ \cline{2-2} \cline{5-8}
                         & BCFA                      &                       &
                          & 0.0685          & 0.1763           & 0.6417           & 0.2388                
                         \\ \cline{2-8} 
                         & \multirow{3}{*}{BCFSA}    & No                    & No                          & 0.0710          & 0.1762           & 0.6435           & 0.2404               \\ \cline{3-8} 
                         &                           & Yes                   & No                          & 0.0713          & 0.1779           & 0.6457           & 0.2416               \\ \cline{3-8} 
                         &                           & Yes                   & Yes                         & \textbf{0.0969} & \textbf{0.1839}  & \textbf{0.6537}  & \textbf{0.2579}      \\ \hline
\end{tabular}}
\label{Tab:tsn-rgb}
\end{table}

%%%%%%%%%%%%%%%%%%%%%%%%%%%%%%%%%%%%%%%%%%%%%%%%%%%%%%%%%%%%%%%%%%%%%%%%%%%%%%%%%%%%%%%%%%%%%%%%%%%%%%%%%%

\begin{table}[b!]
\centering
\caption{Performance comparison of various TSN-Flow model configurations on the BoLD validation set.}
\resizebox{8.6cm}{!}{
\begin{tabular}{|c|c|c|c|c|c|c|c|}
\hline
\multirow{2}{*}{Model}    & \multirow{2}{*}{Features} & \multirow{2}{*}{$\mathcal{L}_{\textrm{emb}}$} & \multirow{2}{*}{Partial BN} & Regression      & \multicolumn{2}{c|}{Classification} & \multirow{2}{*}{ERS} \\ \cline{5-7}
                          &                           &                       &                             & m$R^{2}$\mbox{}$\uparrow$              & mAP$\uparrow$              & mRA$\uparrow$              &                      \\ \hline
\multirow{6}{*}{TSN-Flow} & B                         & \multirow{3}{*}{No}   & \multirow{3}{*}{No}         & 0.0560          & 0.1431           & 0.5778           & 0.2082               \\ \cline{2-2} \cline{5-8} 
                          & BC                        &                       &                             & 0.0661          & 0.1415           & 0.5882           & 0.2155               \\ \cline{2-2} \cline{5-8} 
                          & BF                        &                       &                             & 0.0649          & 0.1497           & 0.5971           & 0.2190               \\ \cline{2-8} 
                          & \multirow{3}{*}{BCF}      & No                    & No                          & 0.0795          & 0.1524           & 0.6054           & 0.2292               \\ \cline{3-8} 
                          &                           & Yes                   & No                          & 0.0888          & \textbf{0.1563}  & 0.6135           & 0.2369               \\ \cline{3-8} 
                          &                           & Yes                   & Yes                         & \textbf{0.0947} & 0.1554           & \textbf{0.6149}  & \textbf{0.2398}      \\ \hline
\end{tabular}}
\label{Tab:tsn-flow}
\end{table}

%%%%%%%%%%%%%%%%%%%%%%%%%%%%%%%%%%%%%%%%%%%%%%%%%%%%%%%%%%%%%%%%%%%%%%%%%%%%%%%%%%%%%%%%%%%%%%%%%%%%%%%%%%

\subsection{Ablation Studies}
\label{ssec:ablation}

Tables \ref{Tab:tsn-rgb} and \ref{Tab:tsn-flow} present performance comparisons among all TSN-RGB and TSN-Flow configurations, respectively, which we have considered. The second columns describe the various input streams that are being included, with \enquote{B} denoting the body, \enquote{C} denoting the context, \enquote{F} denoting the face, \enquote{S} denoting the Places365 scene categories and \enquote{A} denoting the corresponding SUN attributes.

%%%%%%%%%%%%%%%%%%%%%%%%%%%%%%%%%%%%%%%%%%%%%%%%%%%%%%%%%%%%%%%%%%%%%%%%%%%%%%%%%%%%%%%%%%%%%%%%%%%%%%%%%%

\subsubsection{TSN-RGB}

Both the inclusion of the context and face streams are conducive to an increase in ERS score, with the latter showcasing a more considerable boost in performance over the bare-bones body stream. The sole inclusion of either the Places365 scene-specific or the SUN attribute scores is conducive to higher recognition scores, with the latter seemingly being more beneficial. However, it is the combined usage of the two that results in the biggest improvement in performance, in both the categorical and continuous tasks, as expected. Moreover, while the addition of $\mathcal{L}_{\textrm{emb}}$ leads to a trivial performance boost, the application of the $\textit{Partial BN}$ regularization scheme tops off our previously best performing network, reaching a maximum of 0.2579 ERS on the BoLD validation set. It seems as if the continuous re-estimation of mean and variance parameters of batch-normalization layers that are located deeper within the network, becomes obsolete and may in fact have a negative impact on generalization performance, provided that the model's parameters have been previously initialized through a proper pre-training procedure.

The beneficial influence of scene and attribute related features in human emotion understanding becomes more evident in cases where the facial characteristics and poses of the depicted agents are occluded. This is further highlighted in Fig. \ref{fig:scenes_attrib} which includes instances that were randomly selected from the validation set. Each instance is accompanied by its top-5 predicted scene categories, top-9 predicted attributes, ground truth and predicted (regressed) emotion categories (VAD values) as well as the corresponding \textit{Jaccard similarity coefficient} (\textrm{JC}), for each model configuration. Correct category recognition is indicated in green. In all cases, the incorporation of scene and attribute characteristics, on top of the existing bodily, contextual and facial features, results in more emotions being correctly recognised. In addition, emotions that are semantically related, i.e. peace-happiness-pleasure (e) and sadness-suffering-pain (f), are jointly predicted, even though some of them have not been included by the annotators.  

%%%%%%%%%%%%%%%%%%%%%%%%%%%%%%%%%%%%%%%%%%%%%%%%%%%%%%%%%%%%%%%%%%%%%%%%%%%%%%%%%%%%%%%%%%%%%%%%%%%%%%%%%%

\subsubsection{TSN-Flow}

The introduction of either the context or face stream leads to a marginal improvement over the barebones temporal body stream. A more considerable boost in performance is achieved through the inclusion of all three input streams. These findings validate our intuitive decision to follow a multi-stream approach for encoding motion through \textit{Optical Flow}, analogously to the case of the \textit{RGB} modality. As mentioned in the case of TSN-RGB, the addition of the categorical label embedding loss $\mathcal{L}_{\textrm{emb}}$ improves the network's performance in both categorical and continuous tasks, while with the application of the $\textit{Partial BN}$ scheme, the resulting model tops off all previous configurations reaching a maximum of 0.2398 ERS.

%%%%%%%%%%%%%%%%%%%%%%%%%%%%%%%%%%%%%%%%%%%%%%%%%%%%%%%%%%%%%%%%%%%%%%%%%%%%%%%%%%%%%%%%%%%%%%%%%%%%%%%%%%

\begin{table}[t!]
\centering
\caption{Performance comparison of various ST-GCN model configurations on the BoLD validation and test sets.}
\resizebox{8.6cm}{!}{
\begin{tabular}{|c|c|c|c|c|c|c|c|}
\hline
\multirow{2}{*}{Set}    & \multirow{2}{*}{Model}         & \multirow{2}{*}{Pre-training} & \multirow{2}{*}{\begin{tabular}[c]{@{}c@{}}Labeling\\ Strategy\end{tabular}} & Regression      & \multicolumn{2}{c|}{Classification} & \multirow{2}{*}{ERS} \\ \cline{5-7}
                        &                                &                               &                                                                              & m$R^2$\mbox{}$\uparrow$             & mAP$\uparrow$              & mRA$\uparrow$              &                      \\ \hline
\multirow{4}{*}{Valid.} & \multirow{4}{*}{ST-GCN (ours)} & \multirow{3}{*}{None}         & Uniform                                                                      & 0.0245          & 0.1295           & 0.5701           & 0.1871               \\ \cline{4-8} 
                        &                                &                               & Distance                                                                     & 0.0323          & 0.1383           & 0.5841           & 0.1967               \\ \cline{4-8} 
                        &                                &                               & \multirow{2}{*}{Spatial}                                                     & 0.0383          & 0.1387           & 0.5838           & 0.1998               \\ \cline{3-3} \cline{5-8} 
                        &                                & Kinetics {\cite{kay2017kinetics}}               &                                                                              & \textbf{0.0652} & \textbf{0.1542}  & \textbf{0.6103}  & \textbf{0.2237}      \\ \hline\hline
\multirow{2}{*}{Test}   & Luo et al. \cite{Luo2019ARBEETA}              & N/A                           & N/A                                                                          & 0.0440          & 0.1263           & 0.5596           & 0.1940               \\ \cline{2-8} 
                        & ST-GCN (ours)                  & Kinetics \cite{kay2017kinetics}             & Spatial                                                                      & \textbf{0.0908} & \textbf{0.1694}  & \textbf{0.6268}  & \textbf{0.2444}      \\ \hline
\end{tabular}}
\label{Tab:st-gcn}
\vspace{-0.2cm}
\end{table}

%%%%%%%%%%%%%%%%%%%%%%%%%%%%%%%%%%%%%%%%%%%%%%%%%%%%%%%%%%%%%%%%%%%%%%%%%%%%%%%%%%%%%%%%%%%%%%%%%%%%%%%%%%

\subsubsection{ST-GCN}

As far as skeleton-based learning is concerned, Table \ref{Tab:st-gcn} provides a performance comparison among all ST-GCN configurations which we have considered. We notice that the $\textit{spatial}$ labeling strategy leads to better results compared to the others, confirming the findings of \cite{yan2018} but having a relatively minor impact on the overall emotion recognition performance. Pre-training the ST-GCN on Kinetics provides a significant performance boost 
in both categorical and continuous tasks 
over all of its counterparts that have been trained on BoLD from scratch, reaching a maximum ERS of 0.2237 on the validation set and 0.2444 on the test set. Therefore, pre-training plays a crucial role in the performance of the network and presumably constitutes the main differentiating factor between the reported results of \cite{Luo2019ARBEETA}, and ours.

%%%%%%%%%%%%%%%%%%%%%%%%%%%%%%%%%%%%%%%%%%%%%%%%%%%%%%%%%%%%%%%%%%%%%%%%%%%%%%%%%%%%%%%%%%%%%%%%%%%%%%%%%%

\begin{table}[t!]
\centering
\caption{Performance comparison of various network ensembles on the BoLD validation set, utilizing late fusion schemes.}
\resizebox{8.6cm}{!}{
\begin{tabular}{|c|c|c|c|c|c|}
\hline
\multirow{2}{*}{Network Ensembles}                                                  & \multirow{2}{*}{Score Fusion} & Regression      & \multicolumn{2}{c|}{Classification} & \multirow{2}{*}{ERS} \\ \cline{3-5}
                                                                                    &                               & m$R^2$\mbox{}$\uparrow$             & mAP$\uparrow$              & mRA$\uparrow$              &                      \\ \hline
\multirow{2}{*}{\begin{tabular}[c]{@{}c@{}}TSN-RGB+TSN-Flow\\(ours)\end{tabular}}                                                  & Maximum                       & 0.0939          & 0.1840           & 0.6543           & 0.2566               \\ \cline{2-6} 
                                                                                    & Average                       & 0.1444          & 0.1883           & 0.6661           & 0.2858               \\ \hline
\multirow{3}{*}{\begin{tabular}[c]{@{}c@{}}TSN-RGB+TSN-Flow\\+ST-GCN (ours)\end{tabular}} & Maximum                       & 0.0652          & 0.1809           & 0.6493           & 0.2402               \\ \cline{2-6} 
                                                                                    & Average                       & 0.1438          & \textbf{0.1933}  & 0.6658           & 0.2867               \\ \cline{2-6} 
                                                                                    & Weighted Average              & \textbf{0.1489} & 0.1929           & \textbf{0.6682}  & \textbf{0.2897}      \\ \hline
Filntisis et al. \cite{NTUA_BEEU}                                                                    & Average                       & 0.0917          & 0.1656           & 0.6266           & 0.2439               \\ \hline
\end{tabular}}
\label{Tab:proposed}
\vspace{-0.2cm}
\end{table}

%%%%%%%%%%%%%%%%%%%%%%%%%%%%%%%%%%%%%%%%%%%%%%%%%%%%%%%%%%%%%%%%%%%%%%%%%%%%%%%%%%%%%%%%%%%%%%%%%%%%%%%%%%

\begin{figure}[b!]
\centering
\hspace{-0.01cm}\includegraphics[scale=0.553]{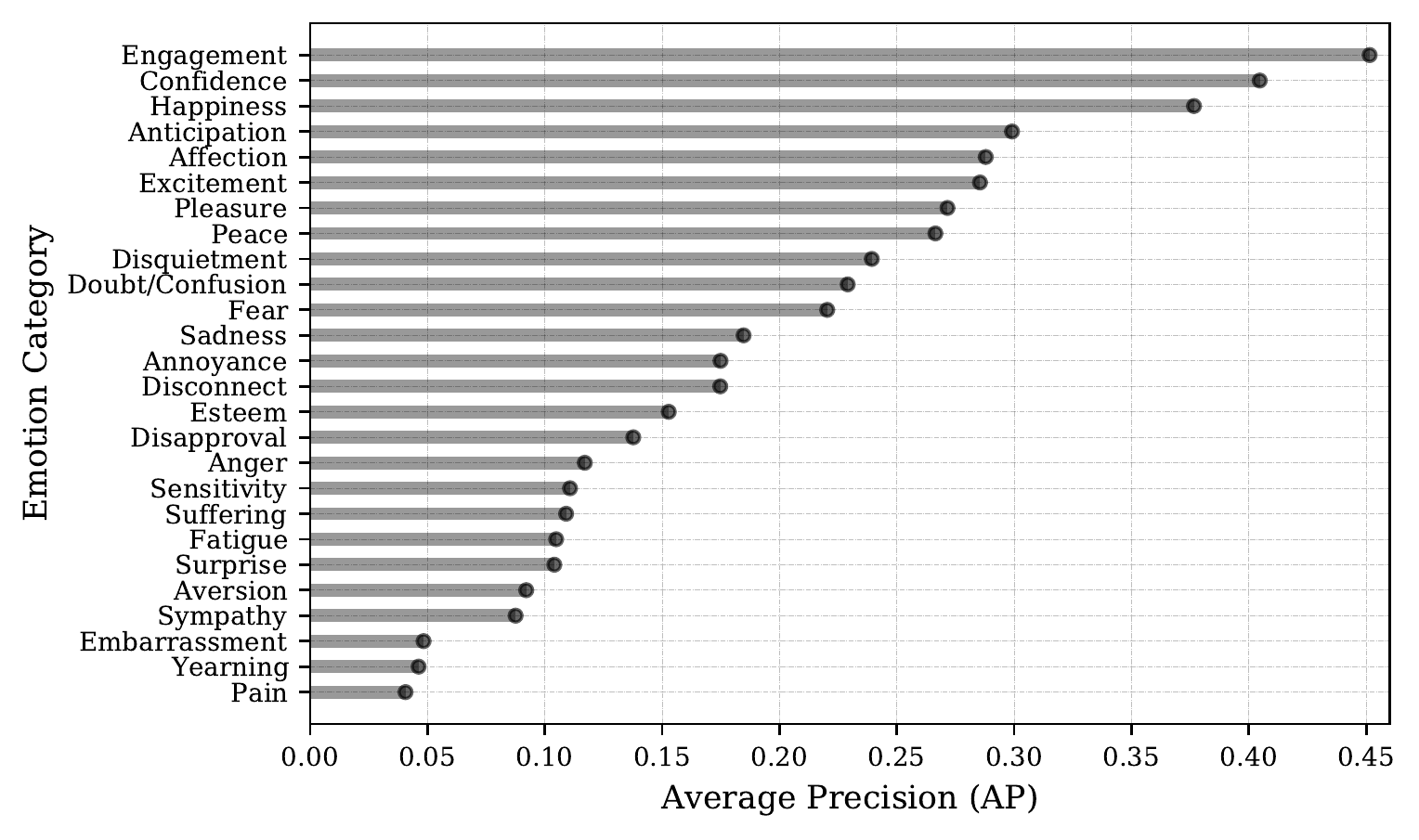}
\caption{\textit{Average precision} (AP) scores per emotion category, as obtained on the BoLD validation set, using our proposed network ensemble.}
\label{fig:val-APs}
\vspace{-0.2cm}
\end{figure}

\begin{figure}[b!]
\hspace{0.425cm}\includegraphics[scale=0.554]{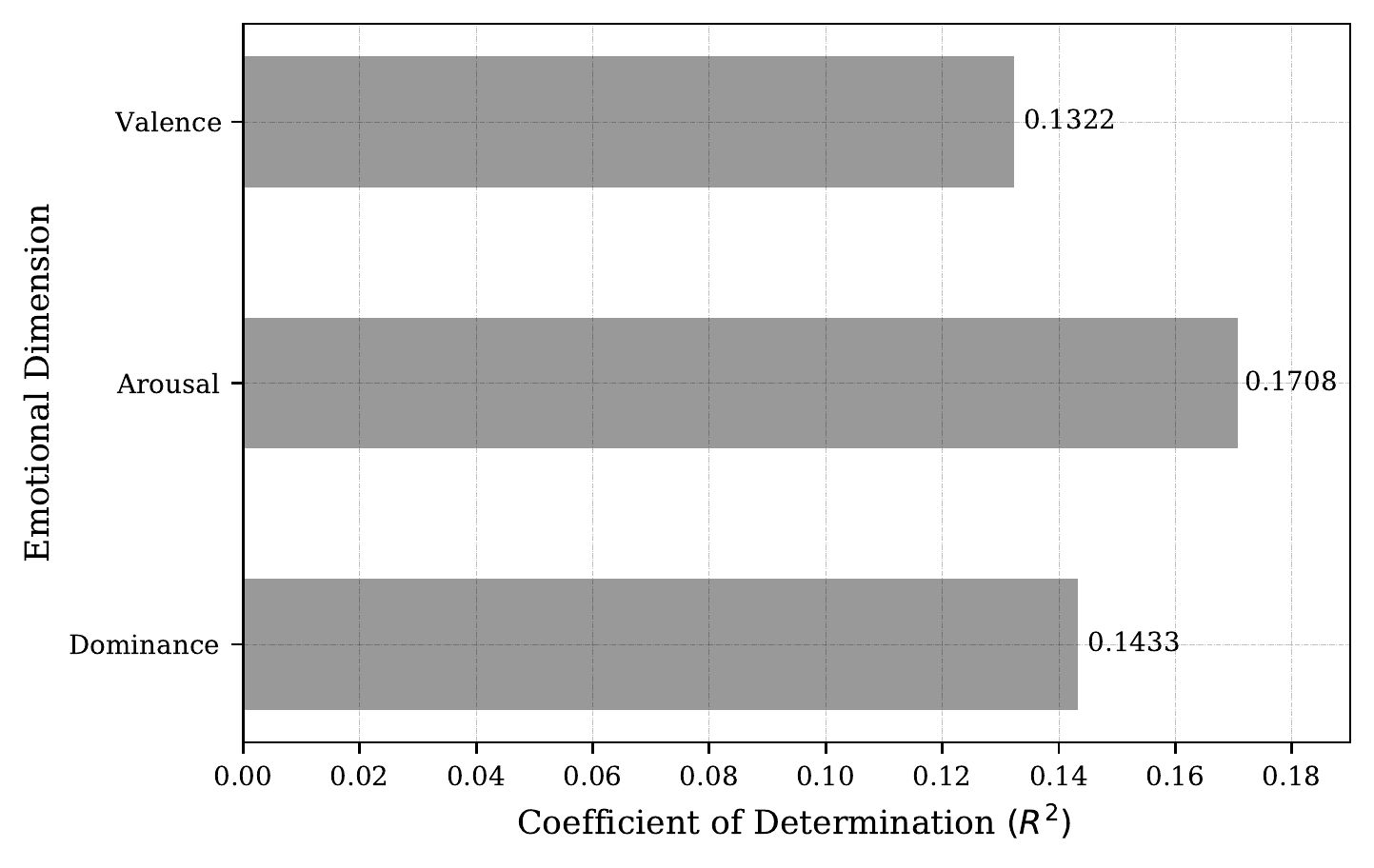}
\caption{\textit{Coefficient of determination} $(R^2)$ per emotional dimension, as obtained on the BoLD validation set, using our proposed network ensemble.}
\label{fig:val-R2s}
\end{figure}

%%%%%%%%%%%%%%%%%%%%%%%%%%%%%%%%%%%%%%%%%%%%%%%%%%%%%%%%%%%%%%%%%%%%%%%%%%%%%%%%%%%%%%%%%%%%%%%%%%%%%%%%%%

\subsubsection{Proposed Method}

The proposed methodology constitutes a late fusion scheme among the best performing models from all modalities, namely \textit{RGB}, \textit{Optical Flow} and \textit{Human Skeleton}, effectively forming a network ensemble. The score fusion methods which will be considered include: maximum, simple average and weighted average. Table \ref{Tab:proposed} summarizes the results. 
%Fusing the predictions of the two separate TSN modules, through simple averaging, results in significant improvements compared to their individual performances, confirming the supplementary roles of appearance and motion in the encoding of emotional behaviors. 
%The maximum operator consistently performs the worst among the three fusion schemes. 
A weighted average of the TSN-RGB, TSN-Flow and ST-GCN, with a weight ratio of 2:2:1 respectively, leads to the best result of 0.2897 ERS on the validation set. More importantly, our implementation surpasses the current state-of-the-art of 0.2439 ERS on the BoLD validation set, as it was recently achieved in \cite{NTUA_BEEU}. Figs. \ref{fig:val-APs} and \ref{fig:val-R2s} summarize the results for the 26 discrete emotion categories and the continuous VAD dimensions relative to the AP and $R^2$ performance metrics, respectively.
%As a detection threshold for each emotion category, we pick the \textit{equal error rate} point, that is the value at which \textit{precision} equals \textit{recall}. %\textcolor{red}{Notice that even our best model is incapable of predicting a single correct emotion for almost half of the samples in the validation set.} 

%%%%%%%%%%%%%%%%%%%%%%%%%%%%%%%%%%%%%%%%%%%%%%%%%%%%%%%%%%%%%%%%%%%%%%%%%%%%%%%%%%%%%%%%%%%%%%%%%%%%%%%%%%

\begin{table}[t!]
\centering
\caption{Quantitative results on the BoLD test set regarding the performance and complexity of our proposed model versus other published works.}
\resizebox{8.6cm}{!}{
\begin{tabular}{|c|c|c|c|c|c|}
\hline
\multirow{2}{*}{Model}  & \multirow{2}{*}{\begin{tabular}[c]{@{}c@{}}\# Parameters\\$(\times 10^6)$\end{tabular}} & Regression & \multicolumn{2}{c|}{Classification} & \multirow{2}{*}{ERS} \\ \cline{3-5}
                        &                                                                               & m$R^2$\mbox{}$\uparrow$        & mAP$\uparrow$              & mRA$\uparrow$              &                      \\ \hline
Luo et al. {\cite{Luo2019ARBEETA}}       & N/A                                                                      & 0.1030     & 0.1714           & 0.6352            & 0.2530                \\ \hline
Filntisis et al. {\cite{NTUA_BEEU}} & $111.5$                                                                     & 0.1141     & 0.1796           & 0.6416           & 0.2624              \\ \hline
Ours                    & $71.4$                                                                    & \textbf{0.1597}     & \textbf{0.2185}          & \textbf{0.6826}           & \textbf{0.3051}          \\ \hline
\end{tabular}}
\label{Tab:BoLD_test}
\vspace{-0.09cm}
\end{table}

%%%%%%%%%%%%%%%%%%%%%%%%%%%%%%%%%%%%%%%%%%%%%%%%%%%%%%%%%%%%%%%%%%%%%%%%%%%%%%%%%%%%%%%%%%%%%%%%%%%%%%%%%%

Subsequently, we evaluate our best performing network ensemble on the official BoLD test set. 
%The predictions for our final submission are produced using $K=25$ segments for both TSN-RGB and TSN-Flow. 
A comparative study regarding the performance and complexity of our proposed model and earlier published works is presented in \mbox{Table \ref{Tab:BoLD_test}}. The proposed network ensemble manages to surpass the current state-of-the-art of 0.2624 ERS, as achieved in \cite{NTUA_BEEU}, by a considerable margin on all metrics, thus verifying the superiority of our technique. As far as complexity is concerned, our model does a good job at maintaining a lower number of trainable parameters through the efficient utilization of multiple input streams and shallow feature extractors (ResNet-18, Wide-ResNet-18), in comparison with previous implementations that made use of deeper ConvNet backbones \cite{NTUA_BEEU} (ResNet-50 \& 101) and disentangled the various input streams all together \cite{Luo2019ARBEETA}.  

%%%%%%%%%%%%%%%%%%%%%%%%%%%%%%%%%%%%%%%%%%%%%%%%%%%%%%%%%%%%%%%%%%%%%%%%%%%%%%%%%%%%%%%%%%%%%%%%%%%%%%%%%%

%\addtolength{\textheight}{-0.1cm} % This command serves to balance the column lengths

\section{CONCLUSIONS AND FUTURE WORK}
\label{sec:conclusions-futurework}

This study employs two major components of action recognition related literature, namely Temporal Segment Networks (TSN) and Spatial-Temporal Graph Convolutional Networks (ST-GCN) with the aim of extending the concept of context-based emotion recognition in the dynamic setting of video sequences. The most notable contribution of this paper is the extension of the original TSN architecture with the inclusion of multiple input streams that effectively encode bodily, contextual, facial and generic scene-related features, enhancing our model's perception of visual context and emotion in general. 
Our intuitive modifications regarding the incorporation of scene and attribute classification scores, as well as multi-stream optical flow, combined with a properly pre-trained ST-GCN, have led to significant improvements over the state-of-the-art techniques with relation to the \mbox{challenging} Body Language Dataset (BoLD).  
%As we utilize shallow ConvNet feature extractors, the achieved performance boost comes with no additional increase in computational cost relative to all previous published implementations. 

Lastly, a possible future research direction might be proposals for further exploitation of the categorical label dependencies that reside within the datasets and may lead to an additional improvement in categorical emotion prediction. Also, the \textit{Depth} modality has been left relatively unexplored on the subject of context-based emotion recognition in videos. Currently, published data with relation to BoLD are quite scarce and there is definitely a lot of room for improvement. However, our existing results undoubtedly prove that we have made significant steps in the right direction. 

%%%%%%%%%%%%%%%%%%%%%%%%%%%%%%%%%%%%%%%%%%%%%%%%%%%%%%%%%%%%%%%%%%%%%%%%%%%%%%%%%%%%%%%%%%%%%%%%%%%%%%%%%%

{\small
\bibliographystyle{ieee}
\bibliography{main}

\begin{thebibliography}{10}\itemsep=-1pt

\bibitem{kapur2005gesture}
A.~{[Asha] Kapur}, A.~{[Ajay] Kapur}, N.~Virji-Babul, G.~Tzanetakis, and P.~F.
  Driessen.
\newblock Gesture-based affective computing on motion capture data.
\newblock In {\em {Int. Conf. Affective Computing and Intelligent Interaction
  (ACII)}}. Springer, 2005.

\bibitem{banziger2012introducing}
T.~B{\"a}nziger, M.~Mortillaro, and K.~R. Scherer.
\newblock Introducing the {G}eneva multimodal expression corpus for
  experimental research on emotion perception.
\newblock {\em Emotion}, 12(5):1161, 2012.

\bibitem{barrett2010context}
L.~F. Barrett and E.~A. Kensinger.
\newblock Context is routinely encoded during emotion perception.
\newblock {\em {Psychological Science}}, 21(4):595--599, 2010.

\bibitem{bhattacharya2020step}
U.~Bhattacharya, T.~Mittal, R.~Chandra, T.~Randhavane, A.~Bera, and D.~Manocha.
\newblock {STEP}: Spatial temporal graph convolutional networks for emotion
  perception from gaits.
\newblock In {\em Proc. AAAI Conf. Artificial Intelligence}, 2020.

\bibitem{castellano2008emotion}
G.~Castellano, L.~Kessous, and G.~Caridakis.
\newblock Emotion recognition through multiple modalities: face, body gesture,
  speech.
\newblock In {\em Affect and Emotion in Human-Computer Interaction}, pages
  92--103. Springer, 2008.

\bibitem{deng2009imagenet}
J.~Deng, W.~Dong, R.~Socher, L.-J. Li, K.~Li, and L.~Fei-Fei.
\newblock Image{N}et: A large-scale hierarchical image database.
\newblock In {\em {2009 IEEE Conf. Computer Vision and Pattern Recognition
  (CVPR)}}. IEEE, 2009.

\bibitem{dudzik2019context}
B.~Dudzik, M.-P. Jansen, F.~Burger, F.~Kaptein, J.~Broekens, D.~Heylen,
  H.~Hung, M.~A. Neerincx, and K.~P. Truong.
\newblock Context in human emotion perception for automatic affect detection: A
  survey of audiovisual databases.
\newblock In {\em {2019 8th Int. Conf. Affective Computing and Intelligent
  Interaction (ACII)}}. IEEE, 2019.

\bibitem{filntisis2019fusing}
P.~P. Filntisis, N.~Efthymiou, P.~Koutras, G.~Potamianos, and P.~Maragos.
\newblock Fusing body posture with facial expressions for joint recognition of
  affect in child--robot interaction.
\newblock {\em {IEEE Robotics and Automation Letters}}, 4(4):4011--4018, 2019.

\bibitem{NTUA_BEEU}
P.~P. Filntisis, N.~Efthymiou, G.~Potamianos, and P.~Maragos.
\newblock Emotion understanding in videos through body, context, and
  visual-semantic embedding loss.
\newblock In {\em {Proc. 16th Eur. Conf. Computer Vision Workshops (ECCVW) -
  Workshop on Bodily Expressed Emotion Understanding (BEEU)}}, 2020.

\bibitem{goodfellow2013challenges}
I.~J. Goodfellow, D.~Erhan, P.~L. Carrier, A.~Courville, M.~Mirza, B.~Hamner,
  W.~Cukierski, Y.~Tang, D.~Thaler, D.-H. Lee, et~al.
\newblock Challenges in representation learning: A report on three machine
  learning contests.
\newblock In {\em {Int. Conf. Neural Inf. Processing}}. Springer, 2013.

\bibitem{gunes2007bi}
H.~Gunes and M.~Piccardi.
\newblock Bi-modal emotion recognition from expressive face and body gestures.
\newblock {\em {Journal of Network and Computer Applications}},
  30(4):1334--1345, 2007.

\bibitem{he2016deep}
K.~He, X.~Zhang, S.~Ren, and J.~Sun.
\newblock Deep residual learning for image recognition.
\newblock In {\em {Proc. IEEE Conf. Computer Vision and Pattern Recognition
  (CVPR)}}, 2016.

\bibitem{jung2015joint}
H.~Jung, S.~Lee, J.~Yim, S.~Park, and J.~Kim.
\newblock Joint fine-tuning in deep neural networks for facial expression
  recognition.
\newblock In {\em Proc. Int. Conf. Computer Vision (ICCV)}, 2015.

\bibitem{karg2010recognition}
M.~Karg, K.~K{\"u}hnlenz, and M.~Buss.
\newblock Recognition of affect based on gait patterns.
\newblock {\em {IEEE Trans. Syst., Man, and Cybern., Part B (Cybernetics)}},
  40(4):1050--1061, 2010.

\bibitem{kay2017kinetics}
W.~Kay, J.~Carreira, K.~Simonyan, B.~Zhang, C.~Hillier, S.~Vijayanarasimhan,
  F.~Viola, T.~Green, T.~Back, P.~Natsev, et~al.
\newblock The {K}inetics human action video dataset.
\newblock {\em arXiv preprint arXiv:1705.06950}, 2017.

\bibitem{kipf2017semisupervised}
T.~N. Kipf and M.~Welling.
\newblock Semi-supervised classification with graph convolutional networks.
\newblock {\em arXiv preprint arXiv:1609.02907}, 2017.

\bibitem{kosti2019context}
R.~Kosti, J.~M. Alvarez, A.~Recasens, and A.~Lapedriza.
\newblock Context based emotion recognition using {EMOTIC} dataset.
\newblock {\em {IEEE Trans. Pattern Analysis and Machine Intelligence
  (TPAMI)}}, 42(11):2755--2766, 2019.

\bibitem{Laban1950TheMO}
R.~Laban and L.~Ullmann.
\newblock {\em The mastery of movement}.
\newblock Boston, Plays, 1971.

\bibitem{lee2019context}
J.~Lee, S.~{[Seungryong] Kim}, S.~{[Sunok] Kim}, J.~Park, and K.~Sohn.
\newblock Context-aware emotion recognition networks.
\newblock In {\em {Proc. Int. Conf. Computer Vision (ICCV)}}, 2019.

\bibitem{Li_2019_CVPR}
M.~Li, S.~Chen, X.~Chen, Y.~Zhang, Y.~Wang, and Q.~Tian.
\newblock Actional-structural graph convolutional networks for skeleton-based
  action recognition.
\newblock In {\em Proc. IEEE/CVF Conf. Computer Vision and Pattern Recognition
  (CVPR)}, 2019.

\bibitem{Liu_2020_CVPR}
Z.~Liu, H.~Zhang, Z.~Chen, Z.~Wang, and W.~Ouyang.
\newblock Disentangling and unifying graph convolutions for skeleton-based
  action recognition.
\newblock In {\em Proc. IEEE/CVF Conf. Computer Vision and Pattern Recognition
  (CVPR)}, 2020.

\bibitem{Luo2019ARBEETA}
Y.~Luo, J.~Ye, R.~B. Adams, J.~Li, M.~G. Newman, and J.~Z. Wang.
\newblock {ARBEE}: Towards automated recognition of bodily expression of
  emotion in the wild.
\newblock {\em {Int. Journal Computer Vision (IJCV)}}, 128:1--25, 2019.

\bibitem{mittal2020emoticon}
T.~Mittal, P.~Guhan, U.~Bhattacharya, R.~Chandra, A.~Bera, and D.~Manocha.
\newblock Emoti{C}on: Context-aware multimodal emotion recognition using
  {F}rege's principle.
\newblock In {\em {Proc. IEEE/CVF Conf. Computer Vision and Pattern Recognition
  (CVPR)}}, 2020.

\bibitem{mobbs2006kuleshov}
D.~Mobbs, N.~Weiskopf, H.~C. Lau, E.~Featherstone, R.~J. Dolan, and C.~D.
  Frith.
\newblock {The Kuleshov Effect: The influence of contextual framing on
  emotional attributions}.
\newblock {\em {Social Cognitive and Affective Neuroscience}}, 1(2):95--106,
  2006.

\bibitem{mollahosseini2017affectnet}
A.~Mollahosseini, B.~Hasani, and M.~H. Mahoor.
\newblock Affect{N}et: A database for facial expression, valence, and arousal
  computing in the wild.
\newblock {\em {IEEE Trans. Affective Computing}}, 10(1):18--31, 2017.

\bibitem{patterson2012sun}
G.~Patterson and J.~Hays.
\newblock {SUN attribute database: Discovering, annotating, and recognizing
  scene attributes}.
\newblock In {\em 2012 IEEE Conf. Computer Vision and Pattern Recognition
  (CVPR)}. IEEE, 2012.

\bibitem{pennington2014glove}
J.~Pennington, R.~Socher, and C.~D. Manning.
\newblock Glo{V}e: Global vectors for word representation.
\newblock In {\em {Proc. 2014 Conf. Empirical Methods Natural Language
  Processing (EMNLP)}}, 2014.

\bibitem{piana2013set}
S.~Piana, A.~Staglian{\`o}, A.~Camurri, and F.~Odone.
\newblock A set of full-body movement features for emotion recognition to help
  children affected by autism spectrum condition.
\newblock In {\em IDGEI Int. Workshop}, 2013.

\bibitem{righart2008recognition}
R.~Righart and B.~De~Gelder.
\newblock Recognition of facial expressions is influenced by emotional scene
  gist.
\newblock {\em {Cognitive, Affective, \& Behavioral Neuroscience}},
  8(3):264--272, 2008.

\bibitem{mehrabianrussell77}
J.~Russell and A.~Mehrabian.
\newblock Evidence for a three-factor theory of emotions.
\newblock {\em {Journal of Research in Personality}}, 11:273--294, 1977.

\bibitem{saha2014study}
S.~Saha, S.~Datta, A.~Konar, and R.~Janarthanan.
\newblock A study on emotion recognition from body gestures using {Kinect}
  sensor.
\newblock In {\em 2014 Int. Conf. Communication and Signal Processing (ICCSP)}.
  IEEE, 2014.

\bibitem{sapinski2019emotion}
T.~Sapi{\'n}ski, D.~Kami{\'n}ska, A.~Pelikant, and G.~Anbarjafari.
\newblock Emotion recognition from skeletal movements.
\newblock {\em {Entropy}}, 21(7):646, 2019.

\bibitem{sheng2021multi}
W.~Sheng and X.~Li.
\newblock Multi-task learning for gait-based identity recognition and emotion
  recognition using attention enhanced temporal graph convolutional network.
\newblock {\em Pattern Recognition}, 114, 2021.

\bibitem{Shi_2019_CVPR}
L.~Shi, Y.~Zhang, J.~Cheng, and H.~Lu.
\newblock Two-stream adaptive graph convolutional networks for skeleton-based
  action recognition.
\newblock In {\em Proc. IEEE/CVF Conf. Computer Vision and Pattern Recognition
  (CVPR)}, 2019.

\bibitem{Si_2019_CVPR}
C.~Si, W.~Chen, W.~Wang, L.~Wang, and T.~Tan.
\newblock An attention enhanced graph convolutional lstm network for
  skeleton-based action recognition.
\newblock In {\em Proc. IEEE/CVF Conf. Computer Vision and Pattern Recognition
  (CVPR)}, 2019.

\bibitem{simonyan2014two}
K.~Simonyan and A.~Zisserman.
\newblock Two-stream convolutional networks for action recognition in videos.
\newblock In {\em {Advances Neural Inf. Processing Syst. (NIPS)}}, pages
  568--576, 2014.

\bibitem{susskind2010toronto}
J.~M. Susskind, A.~K. Anderson, and G.~E. Hinton.
\newblock The {T}oronto face database.
\newblock {\em {Dept. of Computer Science, Univ. of Toronto, Toronto, ON,
  Canada, Tech. Rep}}, 3, 2010.

\bibitem{wang2016temporal}
L.~Wang, Y.~Xiong, Z.~Wang, Y.~Qiao, D.~Lin, X.~Tang, and L.~Van~Gool.
\newblock Temporal {S}egment {N}etworks: Towards good practices for deep action
  recognition.
\newblock In {\em {Eur. Conf. Computer Vision (ECCV)}}. Springer, 2016.

\bibitem{wieserbrosch}
M.~J. Wieser and T.~Brosch.
\newblock Faces in context: A review and systematization of contextual
  influences on affective face processing.
\newblock {\em {Frontiers in Psychology}}, 3:471, 2012.

\bibitem{xiao2010sun}
J.~Xiao, J.~Hays, K.~A. Ehinger, A.~Oliva, and A.~Torralba.
\newblock {SUN} database: Large-scale scene recognition from abbey to zoo.
\newblock In {\em {2010 IEEE Computer Society Conf. Computer Vision and Pattern
  Recognition (CVPR)}}. IEEE, 2010.

\bibitem{yan2018}
S.~Yan, Y.~Xiong, and D.~Lin.
\newblock Spatial temporal graph convolutional networks for skeleton-based
  action recognition.
\newblock In {\em {AAAI}}, pages 7444--7452. {AAAI} Press, 2018.

\bibitem{zach2007duality}
C.~Zach, T.~Pock, and H.~Bischof.
\newblock A duality based approach for realtime {TV-$L^{1}$} optical flow.
\newblock In {\em 29th DAGM Symp. Pattern Recognition}. Springer, 2007.

\bibitem{zagoruyko2016wide}
S.~Zagoruyko and N.~Komodakis.
\newblock Wide residual networks.
\newblock {\em arXiv preprint arXiv:1605.07146}, 2016.

\bibitem{zhang2017facial}
K.~Zhang, Y.~Huang, Y.~Du, and L.~Wang.
\newblock Facial expression recognition based on deep evolutional
  spatial-temporal networks.
\newblock {\em IEEE Trans. Image Processing}, 26(9):4193--4203, 2017.

\bibitem{zhou2016places}
B.~Zhou, A.~Khosla, A.~Lapedriza, A.~Torralba, and A.~Oliva.
\newblock Places: An image database for deep scene understanding.
\newblock {\em arXiv preprint arXiv:1610.02055}, 2016.

\end{thebibliography}
}

\end{document}